\documentclass[letterpaper]{article} 
\usepackage{aaai2026}  
\usepackage{times}  
\usepackage{helvet}  
\usepackage{courier}  
\usepackage[hyphens]{url}  
\usepackage{graphicx} 
\usepackage{natbib}  
\usepackage{caption} 
\usepackage{algorithm}
\usepackage{algorithmic}
\usepackage{amsmath,amssymb,amsfonts}
\usepackage{booktabs}       
\usepackage{multirow}       
\usepackage{threeparttable}
\usepackage{subcaption}

\usepackage{newfloat}
\usepackage{listings}
\DeclareCaptionStyle{ruled}{labelfont=normalfont,labelsep=colon,strut=off} 
\floatstyle{ruled}
\newfloat{listing}{tb}{lst}{}
\floatname{listing}{Listing}
\title{Intention-Guided Cognitive Reasoning for Egocentric \\Long-Term Action Anticipation}

\author{
    Qiaohui Chu\textsuperscript{\rm 1,2},
    Haoyu Zhang\textsuperscript{\rm 1,2},
    Meng Liu\textsuperscript{\rm 3$^*$},
    Yisen Feng\textsuperscript{\rm 1},
    Haoxiang Shi\textsuperscript{\rm 1,2},
    Liqiang Nie\textsuperscript{\rm 1\thanks{Corresponding authors.}}
}

\affiliations{
    \textsuperscript{\rm 1}Harbin Institute of Technology (Shenzhen)\quad
    \textsuperscript{\rm 2}Pengcheng Laboratory\quad
    \textsuperscript{\rm 3}Shandong Jianzhu University\\
    {\{qiaohuichu8599, zhang.hy.2019, mengliu.sdu, yisenfeng.hit, shihaoxiang1999, nieliqiang\}@gmail.com}\\
}

\usepackage{bibentry}

\begin{document}

\maketitle

\begin{abstract}
Long-term action anticipation from egocentric video is critical for applications such as human-computer interaction and assistive technologies, where anticipating user intent enables proactive and context-aware AI assistance. However, existing approaches suffer from three key limitations: 1) underutilization of fine-grained visual cues from hand-object interactions, 2) neglect of semantic dependencies between verbs and nouns, and 3) lack of explicit cognitive reasoning, limiting generalization and long-term forecasting ability. To overcome these challenges, we propose INSIGHT, a unified two-stage framework for egocentric action anticipation. In the first stage, INSIGHT focuses on extracting semantically rich features from hand-object interaction regions and enhances action representations using a verb-noun co-occurrence matrix. In the second stage, it introduces a reinforcement learning-based module that simulates explicit cognitive reasoning through a structured process: \textit{visual perception (think)} $\rightarrow$ \textit{intention inference (reason)} $\rightarrow$ \textit{action anticipation (answer)}. 
Extensive experiments on Ego4D, EPIC-Kitchens-55, and EGTEA Gaze+ benchmarks show that INSIGHT achieves state-of-the-art performance, demonstrating its effectiveness and strong generalization capability.
\end{abstract}

\begin{links}
    \link{Code}{https://github.com/CorrineQiu/INSIGHT}
\end{links}

\section{Introduction}
In real-world applications such as human-computer interaction~\cite{Azam_2024_CVPR,plizzari2024outlook}, augmented reality~\cite{abreu2024parse,xu2024improving}, and assistive systems for visually impaired individuals~\cite{lee2024cookar,xiao2025egoblind}, AI agents must accurately interpret user intent and demonstrate effective long-term planning capabilities within egocentric vision scenarios. The ability to anticipate user actions well in advance allows AI systems to proactively adapt their behaviors, offer timely and contextually appropriate assistance, and avoid potential hazards or inefficiencies. Conversely, insufficient long-term anticipation can result in delayed or inaccurate responses, significantly impacting system reliability and user satisfaction.

To address these critical needs, the \textbf{Long-Term Action Anticipation (LTA)} task has been introduced. LTA focuses on predicting sequences of future actions likely to be executed by individuals wearing first-person cameras, facilitating more proactive and context-aware assistance.  
\begin{figure}[t]
  \centering
  \includegraphics[width=\columnwidth]{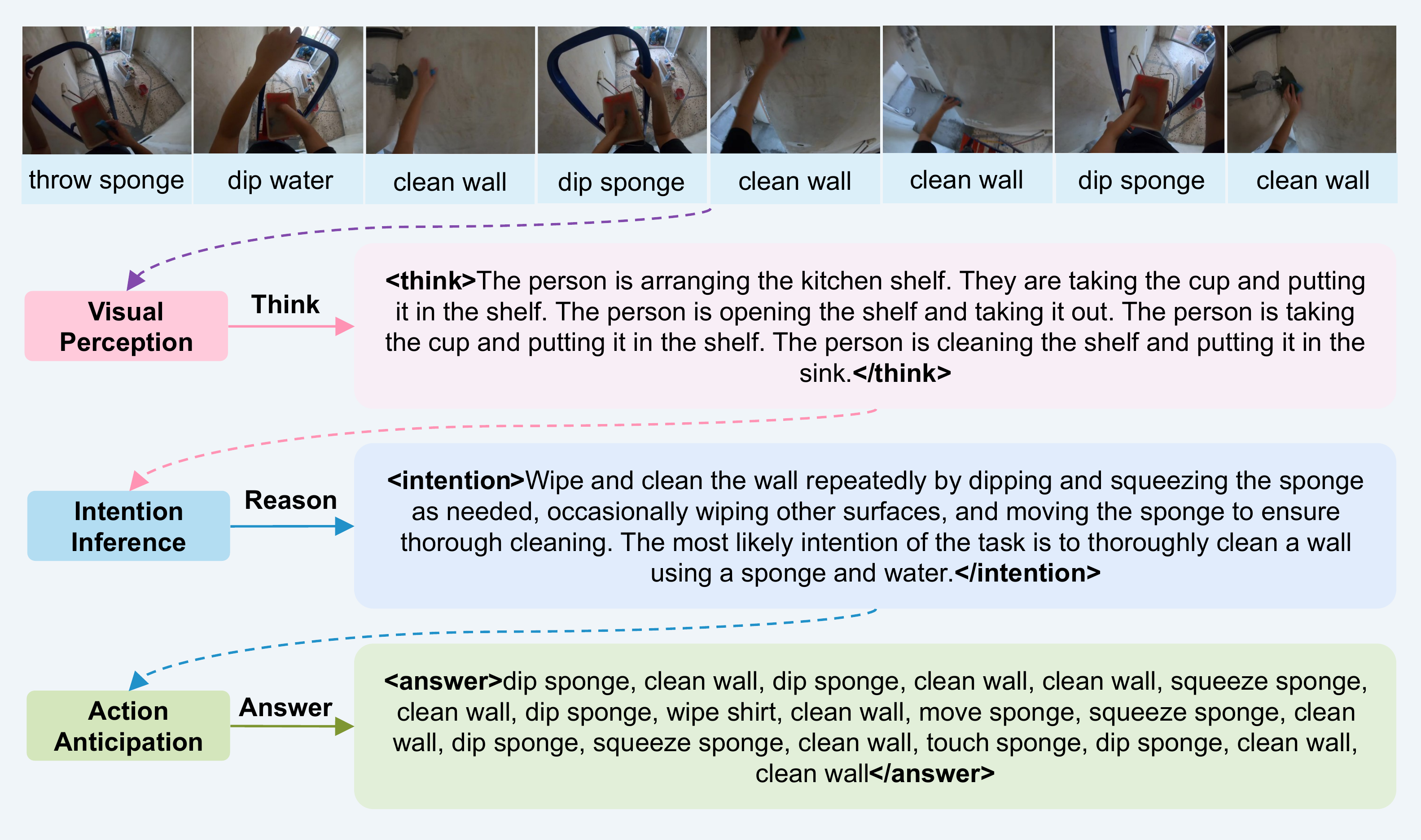}
  \caption{Illustration of the explicit cognitive reasoning process for long-term action anticipation.}

  \label{fig:Reasoning_1}
\end{figure}
Existing approaches to LTA can be broadly categorized into three categories: 1) \textit{Lightweight Temporal Models}~\cite{mascaro2022intention,ashutosh2023hiervl,zhong2024querymamba}. While computationally efficient, they suffer from limited model capacity and inadequate integration of prior knowledge, resulting in poor generalization. 2) \textit{Large Language Models (LLMs)-based Methods}~\cite{chen2023videollm,huang2023palm,mittal2024can}. These methods extract visual content and generate textual descriptions processed by action recognition models or, optionally, Vision-Language Models (VLMs). However, their purely text-based pipeline lacks direct visual grounding, limiting fine-grained scene comprehension and predictive accuracy. 
3) \textit{Action-Embedded Visual Representation Methods}~\cite{cao2025vision}. These approaches incorporate action-related visual embeddings but neglect important egocentric cues and the co-occurrence of verb-noun pairs, both of which are critical for context-aware anticipation. 
Additionally, existing methods predominantly treat LTA as passive sequence prediction, lacking active cognitive reasoning. This limits their robustness and adaptability in dynamic environments.

To address limitations observed in prior research, we propose a unified two-stage framework named INSIGHT, which integrates \textbf{Hand-Object Semantic Action Recognition} with \textbf{Explicit Cognitive Reasoning for Anticipation}. In the first stage, we introduce a specialized feature extraction approach targeting Hand-Object Interaction (HOI) regions to capture crucial egocentric cues effectively. We further enhance contextual understanding by constructing a semantic co-occurrence transition matrix that explicitly models relationships between verbs and nouns. The resulting enriched visual features are then translated into textual prompts and integrated into the VLM to facilitate robust cross-modal representation. 
In the second stage, we present an innovative reinforcement learning-based intention inference module designed for explicit cognitive reasoning. This module simulates structured, human-like thought processes, i.e., \textbf{visual perception (think) \(\rightarrow\) intention inference (reason) \(\rightarrow\) action anticipation (answer)}, as illustrated in Figure~\ref{fig:Reasoning_1}. Unlike passive methods, INSIGHT proactively adapts its predictions by leveraging both observed contexts and inferred user intentions. Extensive evaluations on Ego4D, EPIC-Kitchens-55, and EGTEA Gaze+ demonstrate that INSIGHT significantly surpasses existing state-of-the-art methods.

The main contributions of our work are as follows:
\textbf{1)} We propose a hand-object semantic action recognition module that captures fine-grained HOI cues and incorporates semantic co-occurrence correction, resulting in more robust visual representations.
\textbf{2)} We introduce a reinforcement learning-based intention reasoning mechanism, enabling explicit, human-like cognitive reasoning for LTA.
\textbf{3)} Extensive experiments across multiple benchmarks demonstrate that INSIGHT consistently outperforms state-of-the-art approaches, validating its effectiveness and generalizability.


\section{Related Work}

\subsubsection{Long-Term Action Anticipation.}
Grauman et al.~\cite{grauman2022ego4d} introduced a foundational end-to-end baseline for long-term temporal prediction, employing SlowFast as the visual encoder coupled with a Transformer for sequence modeling. This approach has inspired numerous subsequent studies, which generally fall into three categories based on modality handling and learning methodologies.

The first category emphasizes enhancing visual modeling through improved feature extraction, aggregation strategies, and advanced spatiotemporal architectures. For instance, lightweight models~\cite{hussein2019timeception,das2022video+,nawhal2022rethinking,ishibashi2023technical} have been proposed to boost efficiency and representation quality. Additionally, Zhang et al.~\cite{zhang2024object} introduced object-centric feature extraction using natural language queries, combined with dual-Transformer modules for encoding and temporal aggregation. Despite these innovations, such methods typically suffer from limited  capacity and an absence of external priors, resulting in poor generalization in complex, dynamic, and cluttered egocentric scenarios.

The second category leverages LLMs for temporal reasoning based on textual inputs derived from action recognition models. For instance, Zhao et al.~\cite{zhao2023antgpt}, Pei et al.~\cite{pei2024egovideo}, and Chu et al.~\cite{chu2025technical} fine-tuned LLMs such as LLaMA2-7B~\cite{touvron2023llama} and Vicuna-7B~\cite{zheng2023judging}, using predicted action sequences as input. Kim et al.~\cite{kim2024palm} further enhanced this setup by incorporating VLM-generated captions to enrich contextual understanding. Nonetheless, these approaches rely exclusively on text-based reasoning, neglect direct visual grounding, and consequently compromising fine-grained scene comprehension,  reducing accuracy in extended future predictions.

The third category aims to integrate visual semantics with LLM  reasoning to enhance predictive accuracy. For example, Cao et al.~\cite{cao2025vision} used VLMs to infer task intentions, subsequently employing these inferred intentions as attention queries within visual backbones for improved intention-aware feature aggregation. Although this hybrid method offers better semantic alignment, it overlooks critical egocentric details such as HOI regions and fails to capture the statistical co-occurrences between verbs and nouns, both essential elements for precise anticipation.

\subsubsection{Large Language Model Reasoning.}

Existing LTA methods leveraging fine-tuned LLMs typically adopt supervised fine-tuning strategies~\cite{lester2021power,liu2021p}, relying on static priors-such as VLM-derived textual intentions or captions~\cite{zhao2023antgpt,kim2024palm}. However, these approaches lack dynamic intention inference capabilities, rendering them fragile in complex, extended temporal contexts. While Generalized Reinforcement Preference Optimization (GRPO)~\cite{shao2024deepseekmath,guo2025deepseek} offers a reward-driven alternative, its generic design fails to incorporate task-specific structure, reasoning constraints, and fine-grained success metrics, thereby limiting its effectiveness for LTA. To overcome this limitation, we propose adapting GRPO's reward function to explicitly address the unique challenges and demands inherent to long-term action anticipation.

\begin{figure*}[t]
  \centering
  \includegraphics[width=\linewidth]{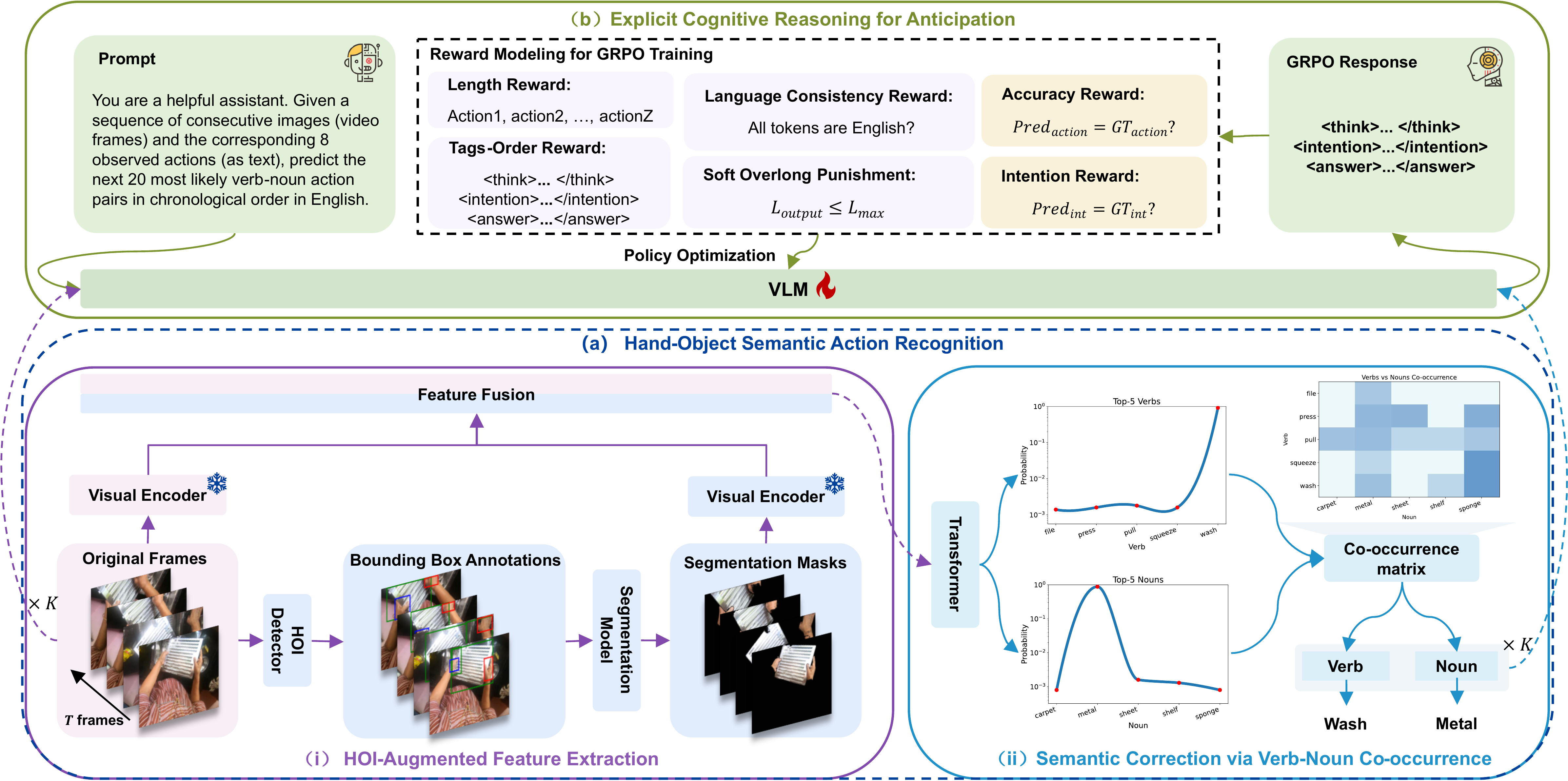}
  \caption{Overview of our two-stage framework, \textbf{INSIGHT}. 
\textbf{Stage 1: Hand-Object Semantic Action Recognition} leverages a HOI-augmented feature extraction module to focus on critical hand-object regions, alongside a semantic co-occurrence transition matrix that captures verb-noun relationships. The resulting enriched visual features are formatted as prompts for VLM. 
\textbf{Stage 2: Explicit Cognitive Reasoning for Anticipation} introduces a reinforcement learning-based intention inference mechanism that simulates a structured three-step cognitive process, i.e., visual perception (\emph{think}), intention inference (\emph{reason}), and action anticipation (\emph{answer}), to enable dynamic reasoning and generate accurate long-term predictions.}

  \label{fig:framework}
\end{figure*}


\section{Method}
An overview of INSIGHT is illustrated in Figure~\ref{fig:framework}, comprising two key stages: hand-object semantic action recognition and explicit cognitive reasoning for anticipation.

\subsection{Problem Definition}
The goal of LTA is to predict future action sequences based on observed egocentric video segments. Given an observed sequence $\mathcal{V}_{obs} = \{\,S_k \mid k = 1,2,\dots,K\,\},$ where $S_k$ represents the $k$-th video segment, the task is to predict the subsequent $Z$ actions, each composed of a verb-noun pair:
\begin{equation}
\bigl\{(\hat v_k, \hat n_k)\bigr\}_{k=K+1}^{K+Z}
= f_{\theta}(\mathcal{V}_{obs}),
\end{equation}
where \((\hat v_k, \hat n_k)\) denotes the predicted verb-noun pair, and \(f_{\theta}\) denotes our predictive model parameterized by 
\(\theta\).

\subsection{Hand-Object Semantic Action Recognition}
Action recognition (AR) plays a pivotal role in egocentric LTA, as it interprets ongoing actions within observed video sequences and provides critical prompts that guide future action predictions. However, most existing AR models in LTA adopt general-purpose visual encoders coupled with Transformer-based architectures, which often fail to capture the unique demands of egocentric perception. While some recent approaches incorporate co-occurrence matrixes to enforce semantic consistency~\cite{zhong2024querymamba}, they still overlook a key aspect: the fine-grained visual information embedded in HOI regions. These regions are densely populated with action-relevant cues and are essential for distinguishing subtle, context-dependent behaviors, as evidenced by recent findings~\cite{zhang2023helping,xu2023egopca,ohkawa2023assemblyhands,zhang2024multi}.
To overcome these limitations, we propose a novel action recognition framework tailored for egocentric LTA that explicitly focuses on capturing discriminative HOI features and incorporates verb-noun co-occurrence statistics as semantic priors. This dual-focus design enhances both the visual grounding and semantic coherence of action predictions, laying a stronger foundation for accurate and reliable long-term anticipation.

\subsubsection{HOI-Augmented Feature Extraction.} Traditional LTA approaches apply visual encoders (e.g., CLIP~\cite{radford2021learning}, EgoVLP~\cite{lin2022egocentric}, EgoVideo-V~\cite{pei2024egovideo}) directly to entire frames. To improve the capture of relevant visual details in egocentric video, we introduce an HOI-focused feature extraction strategy. Specifically, given a video segment \(S_k \in \mathcal{V}_{\mathrm{obs}}\), we follow AntGPT~\cite{zhao2023antgpt} and uniformly sample $4$ frames per segment, denoted as \(F_{k,T}\) with \(T \in \{1, \ldots, 4\}\). Each frame \(F_{k,T}\) undergoes HOI region detection using the pretrained 100DOH detector~\cite{shan2020understanding}, refined with high-resolution masks from SAM2~\cite{ravi2024sam}, producing precise masks \(R_{k,T}\) representing active HOI regions.



We utilize a dual-stream EgoVideo-V architecture to encode both the full frame \(F_{k,T}\) and and the corresponding HOI region 
\(R_{k,T}\), generating visual embeddings: 
\begin{equation}
\bigl(\mathbf{X}_{k,T}^{ori}\,,\;\mathbf{X}_{k,T}^{mask}\bigr)
= \mathrm{EgoVideo\text{-}V}\bigl(F_{k,T},\,{R}_{k,T}\bigr).
\end{equation}
These embeddings are concatenated and fused through a linear Multi-Layer Perceptron (MLP), blending global scene context and detailed HOI features. The resulting integrated representation is processed by a Transformer-based recognition module, effectively capturing complex spatiotemporal relationships and significantly improving semantic accuracy for verb-noun predictions.

\subsubsection{Semantic Correction via Verb-Noun Co-Occurrence.}
To further enhance semantic consistency in action prediction, we implement a semantic correction module utilizing verb-noun co-occurrence statistics derived from training data. This module operates on the output of a Transformer-based model equipped with two parallel classifiers, one for verbs and one for nouns. However, independently predicted verb-noun pairs can result in improbable combinations (e.g., ``drink + guitar''), diminishing prediction reliability. To mitigate this issue, we construct a verb-noun co-occurrence matrix $\mathbf{C}\in\mathbb{N}^{|\mathcal{V}|\times|\mathcal{N}|}$ from annotated labels: 
\begin{equation}
\mathbf{C}_{v,n} \;=\;
\sum_{k=1}^{K}
\mathbf{1}_{\{\,v_k=v \,\wedge\, n_k=n\,\}},
\end{equation}
where \(v\) and \(n\) index the verb and noun categories in \(\mathcal{V}\) and \(\mathcal{N}\), respectively. $\mathbf{1}_{\{\cdot\}}$ denotes the indicator function. The matrix \(\mathbf{C}\) is normalized row-wise and column-wise to obtain conditional probabilities: 
\begin{equation}
\mathbf{P}_{v,n} =
\begin{cases} 
\mathbf{P}^{(n\mid v)}_{v,n} = \frac{\mathbf{C}_{v,n}}{\sum_{n'}\mathbf{C}_{v,n'}} \;\in\; [0,1]^{|\mathcal{V}|\times|\mathcal{N}|}, \\
\mathbf{P}^{(v\mid n)}_{v,n} = \frac{\mathbf{C}_{v,n}}{\sum_{v'}\mathbf{C}_{v',n}} \;\in\; [0,1]^{|\mathcal{V}|\times|\mathcal{N}|}.
\end{cases}
\end{equation}

Given softmax probabilities $p(v_k)$ and $p(n_k)$ from classifiers, we calculate the corrected joint probability:
\begin{equation}
\tilde{p}(v_k,n_k)
\;=\;
p(v_k)\;\cdot\;p(n_k)\;
\cdot
\frac{1}{2}\!
\bigl(\mathbf{P}^{(n\mid v)}_{v,n} + \mathbf{P}^{(v\mid n)}_{v,n}\bigr).
\end{equation}
The final predicted action is determined through maximum a posteriori estimation:
\begin{equation}
(\hat{v}_k,\,\hat{n}_k)
\;=\;
\underset{(v,n)\in\mathcal{V}\times\mathcal{N}}{\arg\max}\;
\tilde{p}(v_k,n_k).
\end{equation}
This semantic correction significantly enhances the coherence and accuracy of predicted verb-noun pairs by aligning predictions with plausible semantic combinations.

\begin{table*}[t]
  \centering
  \begin{threeparttable}
    \begin{tabular*}{\textwidth}{
      @{\extracolsep\fill}
      l c c c c c c
      @{}
    }
      \toprule
      \multirow{2}{*}{Method} & \multirow{2}{*}{Venue} & \multirow{2}{*}{LLM} & \multirow{2}{*}{Visual Encoder} & \multicolumn{3}{c}{Ego4D-v2 (ED)} \\
      \cmidrule(lr){5-7}
      & & & & Verb\,$\downarrow$ & Noun\,$\downarrow$ & Action\,$\downarrow$ \\
      \midrule
      HAI-PUI              & CVPR’24 & --                      & --            & 0.7721 & 0.6733 & 0.9242 \\
      AntGPT               & ICLR’23  & LLaMA2-7B                      & EgoVLP          & 0.6728 & 0.6755 & 0.8931 \\
      PALM                  & ECCV’24  & LLaMA2-7B  & EgoVLP  & 0.7111     & 0.6465     & 0.8819     \\
      PaMsEgoAI     & CVPR’23 & --                      & SlowFast, CLIP            & 0.6702 & 0.6291 & 0.8753 \\
      EgoVideo             & CVPR’24 & Vicuna-7B & EgoVideo-V\tnote{\dag} & 0.6576 & 0.6264 & 0.8619 \\
      ICVL                                & arXiv’25      & LLaMA3-8B                       & ViT-L/14, BLIP2-OPT-2.7B          & \textbf{\underline{0.6516}} & \underline{0.6194} & \underline{0.8570} \\
      \textbf{INSIGHT}                               & --      & Qwen2.5-VL-7B    & EgoVideo-V    & 0.6643 & \textbf{0.6092} & \textbf{0.8463} \\
      \bottomrule
    \end{tabular*}
    \caption{Performance comparison between INSIGHT and prior state-of-the-art methods on the Ego4D-v2 validation set. † indicates models with visual encoders finetuned on the Ego4D dataset. Underlined values represent the current state-of-the-art, and bold values indicate the best (lowest) ED scores.}
    \label{tab:performance}
  \end{threeparttable}
\end{table*}

\subsection{Explicit Cognitive Reasoning for Anticipation}
Most existing LTA methods adopt a pipeline wherein observed egocentric videos are first translated into textual descriptions using action recognition models, followed by fine-tuning LLMs to predict future actions leveraging semantic reasoning. Although effective, these methods employ implicit reasoning and lack explicit modeling of decision-making processes, limiting generalization in diverse or extended scenarios.

To overcome these limitations, we extend GRPO, a reward-driven framework derived from PPO~\cite{schulman2017proximal}, by introducing task-specific structured reasoning, namely ``think $\rightarrow$ reason $\rightarrow$ answer''.  This encourage active intention inference and adaptive decision-making for robust long-term anticipation. Our reward design integrates structural (format) and semantic (content) considerations to ensure precise, interpretable predictions.

\subsubsection{Format Rewards.} We define format-oriented reward functions to enforce structured, interpretable outputs:

\textbf{1) Length Reward}: A full reward is awarded if the generated verb-noun pair set  \(\mathcal{L}\) matches or exceeds the target length \(Z\):
\begin{equation}
S_{len} =
\begin{cases}
1, & \lvert \mathcal{L}\rvert \ge Z,\\
0, & \text{otherwise}.
\end{cases}
\label{eq:s_len}
\end{equation}

\textbf{2) Tags-Order Reward}: To enforce explicit reasoning stages, outputs must adhere to the structured format 
\texttt{<think>...</think>} for visual perception, 
\texttt{<intention>...</intention>} for intention inference, and
\texttt{<answer>...</answer>} for final action prediction:
\begin{equation}
S_{fmt}=
\begin{cases}
1,&\text{all tags are correct},\\
0,&\text{otherwise}.
\end{cases}
\label{eq:s_fmt}
\end{equation}

\textbf{3) Language Consistency Reward}: Penalizes outputs containing non-English tokens to maintain linguistic consistency:
\begin{equation}
S_{lang}=
\begin{cases}
1,&\text{all tokens are English},\\
0,&\text{otherwise}.
\end{cases}
\label{eq:s_lang}
\end{equation}

\textbf{4) Soft Overlong Punishment}: A soft penalty $R_{Soft}$ discourages overly verbose outputs. If the generated length $L$ falls within $[L_{max}-L_{cache}, L_{max}]$, the penalty decays linearly from $0$ to $-1$. If $L$ exceeds $L_{max}$, the penalty is fixed at $-1$. Here \( L_{max} \) denotes the maximum allowable length and \( L_{cache} \) is the buffer length.



\subsubsection{Content Rewards.} Content-oriented rewards are crafted to enhance action prediction accuracy and alignment of inferred intentions:

\textbf{1) Accuracy Reward}: We adopt edit distance (ED) as per Ego4D protocols~\cite{grauman2022ego4d} to evaluate prediction accuracy. 
Given a predicted sequence \(\mathbf{s}_{pred}\) and a ground truth sequence \(\mathbf{s}_{true}\), we normalize the prediction to the target length \(Z\) as follows: If $s_{pred} \textgreater Z$, truncate the sequence; if $s_{pred} \textless Z$, pad with a special token \(\langle{pad}\rangle\).
The resulting normalized prediction sequence is defined as $\tilde{\mathbf{s}}_{\mathrm{pred}} = (\tilde{s}_1, \tilde{s}_2, \dots, \tilde{s}_Z).$ The edit distance between the normalized prediction and the ground truth sequence is computed as:
\begin{equation}
d_{\mathrm{\textit{ED}}}^Z = \mathrm{ED}\bigl(\tilde{\mathbf{s}}_{pred},\,\mathbf{s}_{true}\bigr),
\end{equation}
where $\mathrm{ED}(\cdot,\cdot)$ denotes the edit distance function. The final accuracy reward is normalized to the range $[0,1]$ as follows:
\begin{equation}
S_{acc} = 1 - \frac{d_{ED}^Z}{\lvert \mathbf{s}_{true}\rvert},
\quad
S_{acc} \in [0,1].
\label{eq:s_cont_edit}
\end{equation}
Following the EGO-TOPO~\cite{nagarajan2020ego} setup on the EPIC-Kitchens-55 (EK-55) and EGTEA Gaze+ (EGTEA) datasets, we construct the accuracy reward using mean Average Precision (mAP).




\textbf{2) Intention Reward}: We hypothesize that explicitly articulating high-level task intentions improves long-term action anticipation. To encourage this behavior, we introduce an intention reward that aligns the model’s generated intention  \(int_{gen}\) with a pseudo-ground-truth intention \(int_{gt}\), automatically produced by GPT-4.1. This provides effective supervision without requiring manual annotation. We compute the cosine similarity between the Sentence-BERT~\cite{reimers2019sentence} embeddings of the two intentions:
\begin{equation}
sim
= \cos\bigl(
    Emb(int_{gen}),\,
    Emb(int_{gt})
  \bigr)\in[-1,1],
\end{equation}
The intention reward is normalized using a scaled sigmoid function: 
\begin{equation}
S_{int}
  = \min\!\Bigl(
        \tfrac{1}{1+\exp[-\gamma(sim-\beta)]}\!
        \bigl/\!
        \tfrac{1}{1+\exp[-\gamma(1-\beta)]},\,
        1
      \Bigr),
\quad
\end{equation}
where \(\beta\) and \(\gamma\) control the function's center and sharpness, respectively.

\subsubsection{Overall Reward.} The overall reward function integrates both \textbf{format rewards} and \textbf{content rewards} by using a set of weight coefficients $\{\omega_i\}$ to jointly ensure structural validity and semantic precision in LTA, which is defined as:
\begin{equation}
R = \omega_{1}\,S_{len}\,R_{task} + \omega_{2}\,R_{Soft},
\label{eq:final_reward_revised}
\end{equation}
where $R_{task}$ is the weighted \emph{task score}, which emphasizes action accuracy while incorporating auxiliary signals from intention alignment and output formatting:
\begin{equation}
R_{task}
  =
  \omega_{3}\,S_{acc}
  + \omega_{4}\,S_{int}
  + \omega_{5}\,S_{lang}
  + \omega_{6}\,S_{fmt}.
\end{equation}

\begingroup
  \setlength{\tabcolsep}{3pt}
  \begin{table*}[htbp]
    \centering
    \begin{threeparttable}
      \begin{tabular*}{\textwidth}{@{\extracolsep\fill}
        l
        c
        c
        c
        c
        c
        c
        c
        @{}}
        \toprule
        \multirow{2}{*}{Method} & \multirow{2}{*}{Venue} &
          \multicolumn{3}{c}{EK-55 (mAP)} &
          \multicolumn{3}{c}{EGTEA (mAP)} \\
        \cmidrule(lr){3-5}\cmidrule(lr){6-8}
        & & ALL\,$\uparrow$ & FREQ\,$\uparrow$ & RARE\,$\uparrow$
            & ALL\,$\uparrow$ & FREQ\,$\uparrow$ & RARE\,$\uparrow$ \\
        \midrule
        Timeception       & CVPR’19  & 35.6 & 55.9 & 26.1 & 74.1 & 79.7 & 59.7 \\
        VideoGraph        & ICCV’19  & 22.5 & 49.4 & 14.0 & 67.7 & 77.1 & 47.2 \\
        EGO-TOPO          & CVPR’20  & 38.0 & 56.9 & 29.2 & 73.5 & 80.7 & 54.7 \\
        Anticipatr        & ECCV’22  & 39.1 & 58.1 & 29.1 & 76.8 & 83.3 & 55.1 \\
        AntGPT            & ICLR’23  & 40.1 & 58.8 & 31.9 & 80.2 & 84.8 & 72.9 \\
        PALM              & ECCV’24  & 40.4 & 59.3 & 30.3 & 80.7 & 85.0 & 73.5 \\
        ICVL              & arXiv’25 & \underline{43.3} & \underline{61.6} & \underline{33.8}
                                   & \underline{81.0} & \underline{85.2} & \underline{73.7} \\
        \textbf{INSIGHT}  & --       & \textbf{45.2}      & \textbf{62.4}      & \textbf{36.0}
                                   & \textbf{81.7}      & \textbf{85.9}      & \textbf{74.4}      \\
        \bottomrule
      \end{tabular*}
      \caption{Performance comparison between INSIGHT and prior state-of-the-art methods on the EK-55 validation set and the EGTEA test set. Underlined values indicate the current state-of-the-art, and bold values highlight the highest mAP scores in each column.}
      \label{tab:lta_results}
    \end{threeparttable}
  \end{table*}
\endgroup

\section{Experiment}
\subsection{Experimental Settings}
\subsubsection{Datasets.}
We evaluated our proposed method using three widely recognized egocentric video benchmarks, each encompassing distinct scenarios and action taxonomies:

\textbf{1) Ego4D}~\cite{grauman2022ego4d} is among the most comprehensive egocentric datasets available, capturing diverse daily-life scenarios. The \textit{v1} release includes $3,670$ hours of footage, with the \textit{Forecasting} subset containing $1,723$ annotated clips ($116$ hours) labeled with $115$ verbs and $478$ nouns. The updated \textit{v2} dataset expands this subset to $3,472$ clips ($243$ hours), annotated with $117$ verbs and $521$ nouns. We employed the official data splits from the \textit{v2} version for our experiments.


\textbf{2) EPIC-Kitchens-55}~\cite{Damen2018EPICKITCHENS} comprises egocentric cooking activities recorded in realistic kitchen settings. This dataset includes $55$ hours of densely annotated videos with $125$ verbs and $352$ nouns. Our experiments utilized standard data splits defined by EGO-TOPO~\cite{nagarajan2020ego} to ensure comparability with previous studies. 

\textbf{3) EGTEA Gaze+}~\cite{li2018eye} consists of $86$ cooking-focused egocentric videos totaling $26$ hours. It provides annotations for $19$ verbs and $51$ nouns. Consistent with established evaluation protocols, we followed the official splits proposed in~\cite{nagarajan2020ego}.

\subsubsection{Evaluation Metrics.} 
We adopted evaluation protocols specific to each dataset, ensuring consistency and comparability with prior works. For the Ego4D dataset, we adhered to the official evaluation approach, which specifies an observed context of $8$ segments and requires predicting $5$ distinct future action sequences, each comprising $20$ verb-noun pairs. We evaluated using edit distance~\cite{damerau1964technique} computed separately for verbs, nouns, and combined actions. The evaluation score is the minimal edit distance achieved between any of the predicted sequences and the ground-truth sequence.



For EK-55 and EGTEA, we followed the multi-label classification evaluation protocol proposed by Nagarajan et al.~\cite{nagarajan2020ego}. Specifically, we reported mAP across varying observation horizons. The model is given the first \(P\%\) of a video as input and tasked with predicting verbs that occur in the remaining \((100-P)\%\). We evaluated at \(P \in \{25, 50, 75\}\), and reported mAP across three action categories: all actions (All), frequently occurring actions (Freq), and rare actions (Rare).


\subsubsection{Implementation Details.} 
\begin{figure}[t]
  \centering
  \includegraphics[width=\columnwidth]{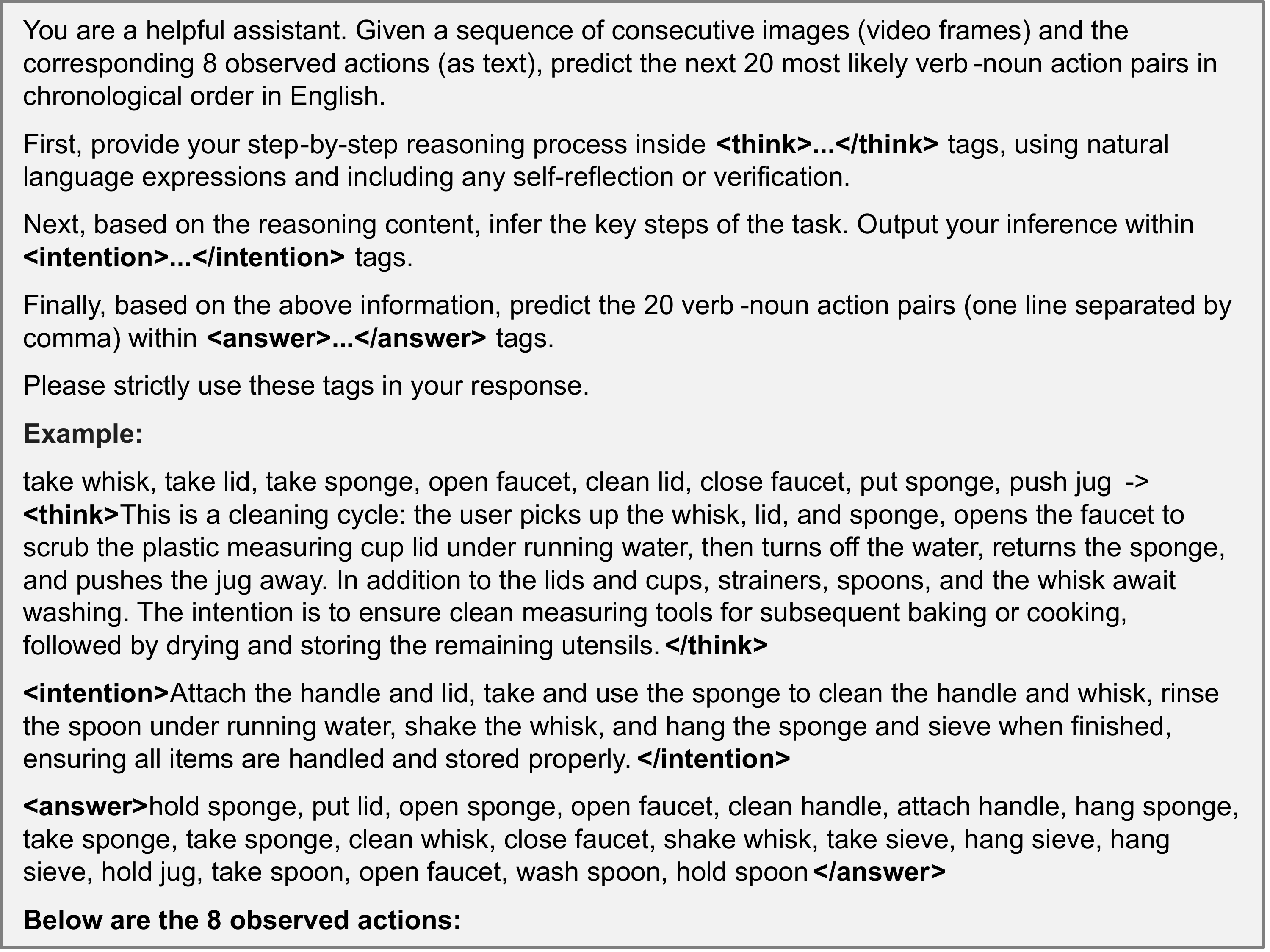}
  \caption{Illustration of the structured prompt used during GRPO training.}

  \label{fig:Prompt}
\end{figure}

We employed a frozen EgoVideo-V encoder to extract visual embeddings from full frames and corresponding HOI regions. These dual-stream features were linearly fused and processed by a Transformer architecture consisting of $4$ layers and $8$ attention heads per layer to capture spatiotemporal dependencies. The feature extraction stage was trained using a batch size of $8$, a learning rate of $8e-5$, and a plateau-based learning rate scheduler. 
For cognitive reasoning, we used Qwen2.5-VL-Instruct-7B~\cite{bai2025qwen2} as the backbone, incorporating structured prompting (Figure~\ref{fig:Prompt}) and trained end-to-end with Swift framework~\cite{zhao2024swiftascalablelightweightinfrastructure} using our GRPO-based reinforcement learning objective. Experiments were conducted on six NVIDIA H20-SXM5-96GB GPUs. 

Hyperparameters for fine-tuning included batch size of $24$, learning rate of $3e-6$, temperature of $0.9$, KL divergence coefficient of $0.08$, generation numbers of $5$, buffer length of $256$, and a maximum allowable length of $450$. Reward function weights were set as follows: $\omega_{1} = 0.90$, $\omega_{2} = 0.10$, $\omega_{3} = 0.85$, $\omega_{4} = 0.05$, $\omega_{5} = 0.05$, and $\omega_{6} = 0.05$. The intention reward parameters were \(\beta=0.8\) and \(\gamma=40\). The full fine-tuning process consists of $500$ steps, taking approximately $90$ GPU hours.

\subsection{Performance Comparison}
\subsubsection{On Ego4D-v2.}
Table~\ref{tab:performance} compares INSIGHT against previous state-of-the-art methods on the Ego4D \textit{v2} validation set. INSIGHT outperforms the best existing method ICVL~\cite{cao2025vision} in noun and combined action predictions by margins of 1.02\% and 1.07\%, respectively. The superior noun prediction was attributed to our HOI-focused feature extraction, capturing essential object manipulations. On ``verb'', our score slightly above the best result. Notably, INSIGHT achieves this slight improvement in verb prediction using a frozen visual encoder, in contrast to EgoVideo~\cite{pei2024egovideo}'s fine-tuned encoder and ICVL's dual-model approach with ViT-L/14 and BLIP2-OPT-2.7B for enhanced visual learning. This suggests that the fine-tuned language model and the proposed cognitive reasoning module effectively compensate for visual ambiguities, leading to more accurate and context-aware long-term predictions.


\subsubsection{On EK-55 and EGTEA.} 
Table~\ref{tab:lta_results} further demonstrates INSIGHT’s superior performance on EK-55 and EGTEA benchmarks, improving on the baseline ICVL by 4.4\%, 1.3\%, and 6.5\% (EK-55) and by 0.9\%, 0.8\%, and 1.0\% (EGTEA) for All, Freq, and Rare actions, respectively. Notably, INSIGHT excels in rare classes, indicating that our cognitive reasoning and intention-alignment mechanisms substantially reduce tail-class confusion, ensuring robust generalization. Moreover, consistent improvements with a reduced head-tail gap showing better calibration, these gains persist across datasets with varying protocols, indicating the structured reasoning trace enhances long-term prediction performance. More experimental results are available in \textbf{Appendix}.


\begin{figure*}[t]
  \centering
  \includegraphics[width=\linewidth]{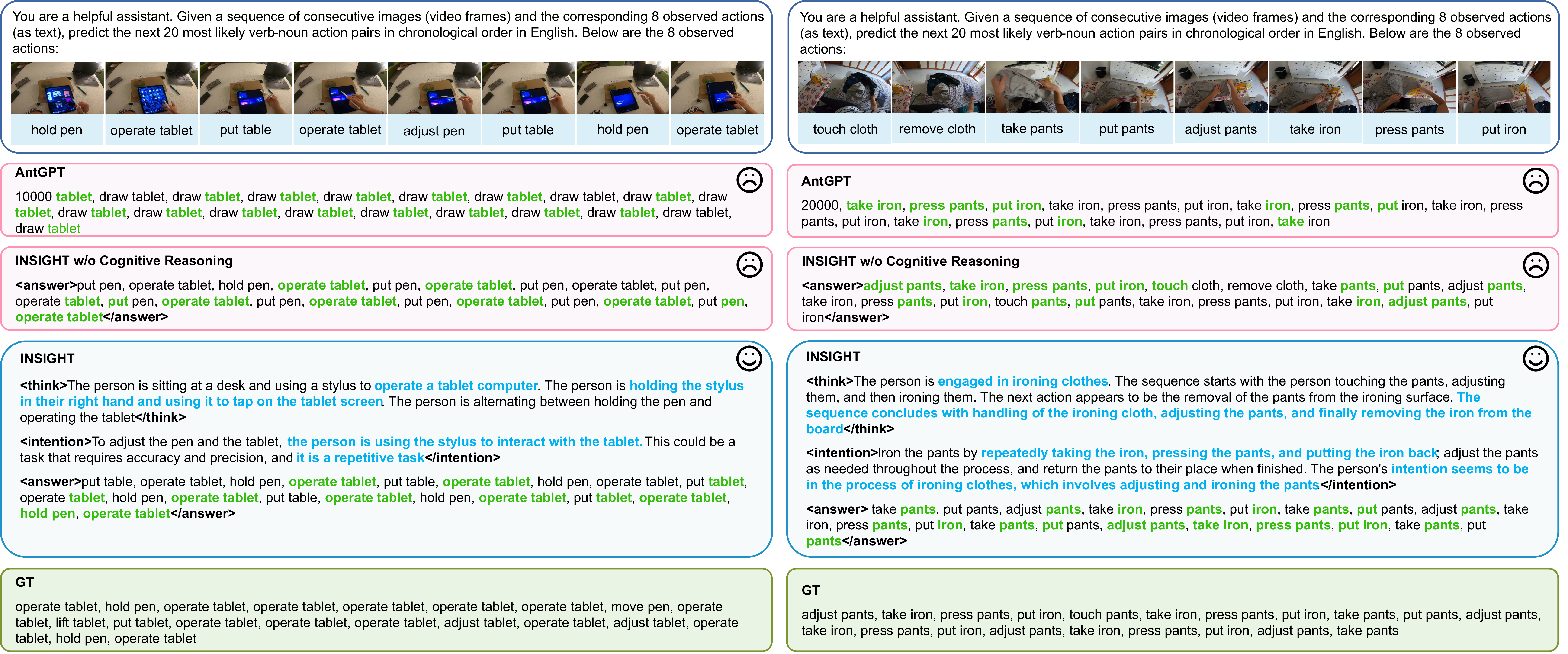}
  \caption{Case study on Ego4D for the LTA task. INSIGHT demonstrates intention-aware predictions with improved temporal coherence, fewer redundant actions, and more precise verb-noun pairings, leading to lower edit distance.}
  \label{fig:result_example}
\end{figure*}

\subsection{Ablation Study}

\begin{table}[t]
  \centering
  \begin{tabular}{lccc}
    \toprule
    \multirow{2}{*}{Method} & \multicolumn{3}{c}{Ego4D-v2 (ED)} \\
    \cmidrule(lr){2-4}
                            & Verb\,$\downarrow$    & Noun\,$\downarrow$    & Action\,$\downarrow$   \\
    \midrule
    w/o HOI feature                 & 0.6719       & 0.6158       & 0.8595        \\
    w/o Semantic correction & 0.6716   & 0.6108       & 0.8587        \\
    w/o Cognitive reasoning                 & 0.6750  & 0.6176  & 0.8612   \\
    w/o Intention           & 0.6685  & 0.6104  & 0.8571   \\
    \midrule
    INSIGHT (full model)       & \textbf{0.6643} & \textbf{0.6092} & \textbf{0.8463} \\
    \bottomrule
  \end{tabular}
  \caption{Ablation study analyzing the contribution of different modules within the INSIGHT framework.}

  \label{tab:ablation}
\end{table}

\begin{figure}[t]
  \centering
  \begin{subfigure}[b]{0.49\columnwidth}
    \includegraphics[width=\linewidth]{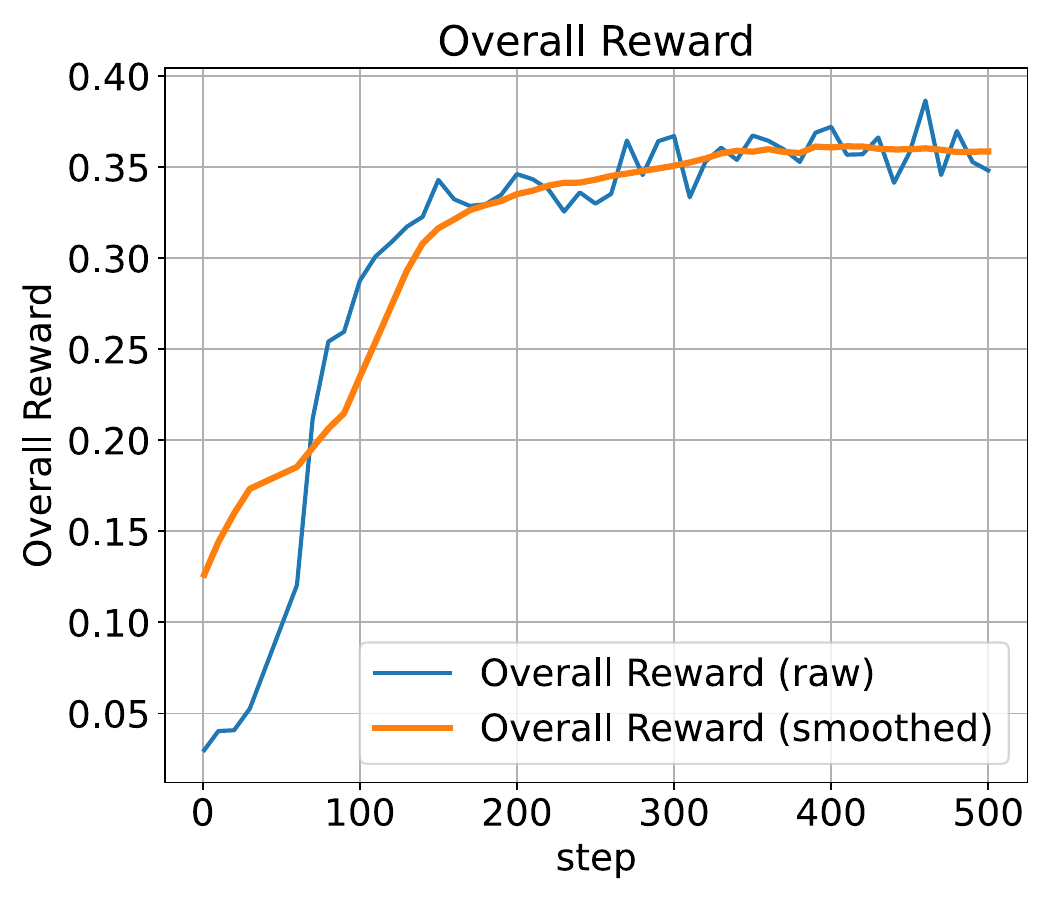}
    \caption{Overall Reward}
    \label{fig:train_ActionReward}
  \end{subfigure}
  \hfill
  \begin{subfigure}[b]{0.49\columnwidth}
    \includegraphics[width=\linewidth]{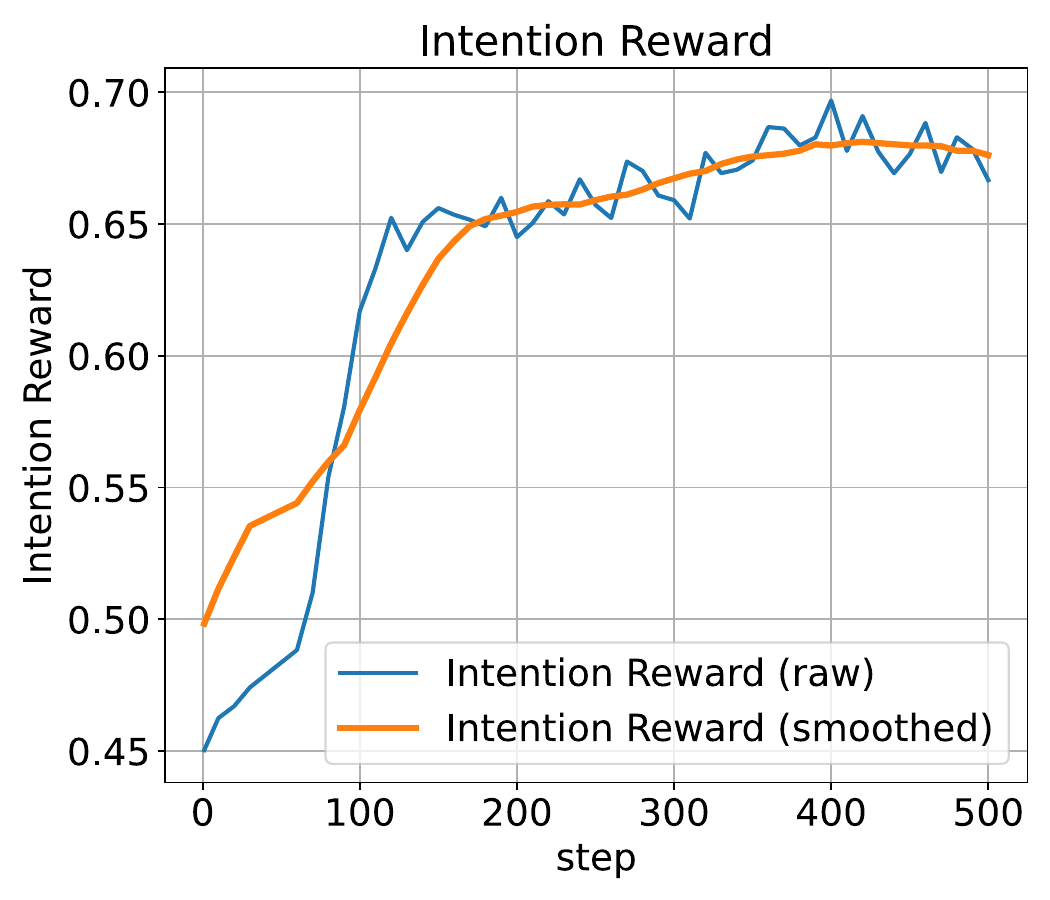}
    \caption{Intention Reward}
    \label{fig:train_IntentRewardRAW}
  \end{subfigure}
  \caption{Training visualization of the GRPO-based cognitive reasoning process.}

  \label{fig:train_rewards}
\end{figure}
We conducted ablation studies to quantify the contribution of each component of INSIGHT. Comparison results are summarized in Table~\ref{tab:ablation}.

\subsubsection{On Action Recognition.}
In the action recognition stage, removing HOI region features and using only raw frames (w/o HOI feature) leads to a clear degradation. This confirms the importance of explicitly modeling HOI cues, which significantly enhances recognition accuracy and supports downstream anticipation. Additionally, removing the verb-noun co-occurrence matrix (w/o Semantic correction) also causes a noticeable drop in performance. This indicates that structured semantic priors contribute to more discriminative and semantically consistent action representations.



\subsubsection{On Anticipation.} In the anticipation stage, discarding the structured reasoning trace (think \(\rightarrow\) reason \(\rightarrow\) answer) and directly predicting future actions (w/o Cognition reasoning) leads to a substantial drop in performance. This demonstrates that stepwise reasoning is crucial for enabling the model to internalize task intentions and temporal dependencies effectively. Furthermore, maintaining the reasoning pipeline but omitting supervision for the intention inference stage (w/o Intention) also results in performance degradation compared to the full model. This highlights the value of supervising intermediate reasoning steps, as it calibrates the cognitive process and improves overall prediction quality. 

Overall, these results show that each module provides complementary gains, and their integration delivers the highest accuracy. Notably, removing the structured reasoning trace (w/o Cognitive reasoning) has the most significant impact on performance, highlighting its importance. These findings align with prior observations that both high-quality action recognition and explicit cognitive reasoning are essential for robust and reliable long-term action anticipation.

\subsection{Reward Analysis}
Figure~\ref{fig:train_rewards} presents learning dynamics under the adopted training configuration. (a) illustrates the progression of the overall reward, while (b) tracks the intention-specific reward; both panels show raw and smoothed curves to capture training trends. Training converges within $500$ steps: rewards rise sharply during the initial $1-100$ steps, followed by gradual improvement and stabilization between $100-300$ steps. Importantly, the intention reward closely mirrors the overall reward trajectory, suggesting that intention supervision is well aligned with the task objective. Although raw curves exhibit fluctuations typical of on-policy learning and exploration, smoothed trends indicate consistent performance gains without signs of instability. These results demonstrate the robustness of GRPO under our selected hyperparameters and validate the utility of explicit intention guidance in improving policy quality and convergence.




\subsection{Qualitative Comparison}
We present a qualitative comparison on Ego4D-v2 by reproducing the AntGPT~\cite{zhao2023antgpt} baseline, which, as shown in Figure~\ref{fig:result_example}, exhibits task-irrelevant tokens, repetitive predictions, and verb-noun mismatches. By contrast, INSIGHT without cognitive reasoning enhances lexical fidelity but suffers from temporal inconsistency, exemplified by repeated ``put pen'' actions without prior ``hold pen''. The full INSIGHT model, equipped with cognitive reasoning, outperforms both by inferring task intention and aligning subgoals, yielding precise verb-noun pairings, expanded verb diversity, and lower edit distances. For further qualitative analyses of hand-object semantic action recognition module, please refer to \textbf{Appendix}.

\section{Conclusion}
This paper introduces INSIGHT, a unified two-stage framework for LTA that combines Hand-Object Semantic Action Recognition with Explicit Cognitive Reasoning. INSIGHT first extracts discriminative visual features from HOI regions and reinforces semantic consistency using a verb-noun co-occurrence prior, enabling the construction of robust action representations. These representations are then processed through a reinforcement learning-based cognitive reasoning module that simulates a structured inference process ``think $\rightarrow$ reason $\rightarrow$ answer'' guided by intention supervision. Extensive experiments on Ego4D, EPIC-Kitchens-55, and EGTEA Gaze+ demonstrate that INSIGHT consistently outperforms prior state-of-the-art methods. As future work, we will capture egocentric dynamics by modeling hand motion trajectories and object state changes in HOI regions as priors, thereby strengthening visual grounding and enhancing long-term anticipation.

\section{Acknowledgments}
This work is supported by Shenzhen Science and Technology Program, No.:KQTD20240729102207002; Jiangsu Science and Technology Major Program, No.:BG2024041; the National Natural Science Foundation of China, No.:62376140 and No.:U23A20315; the Science and Technology Innovation Program for Distinguished Young Scholars of Shandong Province Higher Education Institutions, No.:2023KJ128; the Special Fund for Taishan Scholar Project of Shandong Province; the Major Key Project of Pengcheng Laboratory; the Fundamental Research Funds for the Central Universities, No.:HIT.DZJJ.2025048.

\clearpage
\thispagestyle{empty}

\twocolumn[
  \begin{center}
    {\LARGE\bfseries Intention-Guided Cognitive Reasoning for Egocentric\\Long-Term Action Anticipation}\par
    \vspace{1ex}
    {\large — Supplementary Material —}\par
    \vspace{1ex}

  \end{center}
  \vspace{1em}
]

\section{Grouped Relative Policy Optimization(GRPO)}
We employ the GRPO~\cite{shao2024deepseekmath} algorithm for reinforcement learning. Below, we provide a formal introduction to GRPO, which underpins our policy optimization framework.
Given the input $\boldsymbol{x}$, the policy $\pi_{\theta_{\mathrm{old}}}$ draws a group of $G$ candidate outputs $\mathcal{O}=\{o_j\mid j=1,2,\dots,G\}$. Each output is scored by a task-specific reward function
$r_g\!=\!r(o_g,\boldsymbol{x})$.
The advantage for the $g$-th output is computed by:
\begin{equation}
A_g \;=\;
\frac{\,r_g - \tfrac{1}{G}\sum_{j=1}^{G} r_j\,}
     {\mathrm{std}\!\bigl\{r_1,\dots,r_G\bigr\}},
\quad g=1,\dots,G,
\label{eq:grpo_adv}
\end{equation}
where $\mathrm{std}\!\{\cdot\}$ is the unbiased sample standard deviation.
This eliminates the external value network used in PPO.
Let ${\pi_\theta}$ denote the updated policy.  
The likelihood ratio between new and old policies is:
\begin{equation}
\rho_g
\;=\;
\frac{\pi_\theta(o_g\mid\boldsymbol{x})}
     {\pi_{\theta_{\mathrm{old}}}(o_g\mid\boldsymbol{x})}.
\end{equation}
Following PPO, symmetric clipping with threshold $\varepsilon$
yields the surrogate:
\begin{align}
C(\theta;\boldsymbol{x},\mathcal{O})
&= \frac{1}{G}\sum_{g=1}^{G}
   \min\Bigl(\rho_g A_g, \nonumber\\
&\quad\mathrm{clip}\bigl(\rho_g,1-\varepsilon,1+\varepsilon\bigr)\,A_g\Bigr).
\label{eq:grpo_clip}
\end{align}

A reverse-KL term keeps the new policy close to a reference
SFT model $\pi_{\mathrm{ref}}$:
\begin{align}
\mathcal{J}_{\mathrm{GRPO}}(\theta)
&=
\mathbb{E}_{\boldsymbol{x},\mathcal{O}\sim\pi_\theta}\Bigl[
      C(\theta;\boldsymbol{x},\mathcal{O}) \nonumber\\
&\quad - \beta\,D_{\mathrm{KL}}\bigl[\pi_\theta\parallel\pi_{\mathrm{ref}}\bigr]
\Bigr],
\label{eq:grpo_obj}
\\[4pt]
D_{\mathrm{KL}}\bigl[\pi_\theta\parallel\pi_{\mathrm{ref}}\bigr]
&=
\mathbb{E}_{o\sim\pi_\theta}\Bigl[
      \tfrac{\pi_{\mathrm{ref}}(o\mid\boldsymbol{x})}{\pi_\theta(o\mid\boldsymbol{x})} \nonumber\\
&\quad - \log\!\tfrac{\pi_{\mathrm{ref}}(o\mid\boldsymbol{x})}{\pi_\theta(o\mid\boldsymbol{x})}
      - 1
\Bigr],
\label{eq:grpo_kl}
\end{align}
where $\beta$ controls the strength of the KL regulariser.
The reward function steers policy updates and is pivotal to the stability and effectiveness of reinforcement learning training.


\section{More Experiments}

\subsection{Quantitative Evaluation}

\begin{table*}[htbp]
  \centering
  \begin{threeparttable}
    \begin{tabular*}{\textwidth}{@{\extracolsep{\fill}} l c c c c c c c @{}}
      \toprule
      \multirow{2}{*}{Method}
        & \multirow{2}{*}{Observation Horizon \(P\%\)}
        & \multicolumn{3}{c}{EK-55 (mAP)}
        & \multicolumn{3}{c}{EGTEA (mAP)} \\
      \cmidrule(lr){3-5}\cmidrule(lr){6-8}
      & & ALL $\uparrow$ & FREQ $\uparrow$ & RARE $\uparrow$ & ALL $\uparrow$ & FREQ $\uparrow$ & RARE $\uparrow$ \\
      \midrule
      \multirow{4}{*}{\textbf{INSIGHT}}
        & 25  & 43.6  & 60.9  & 34.7
             & 80.7 & 83.4 & 76.2 \\
        & 50  & 45.9  & 62.6  & 36.5
             & 81.6 & 87.9 & 70.7 \\
        & 75  & 46.1  & 63.7  & 36.7
             & 82.7 & 86.4 & 76.4 \\
        & Avg & \textbf{45.2}  & \textbf{62.4}  & \textbf{36.0}
             & \textbf{81.7} & \textbf{85.9} & \textbf{74.4} \\
      \bottomrule
    \end{tabular*}
    \caption{Performance comparison of INSIGHT and previous state-of-the-art methods on EK-55 validation dataset and EGTEA test dataset, evaluated at three observation horizons \(P \in \{25,50,75\}\). Bold numbers indicate the average across these horizons.}
    \label{tab:lta_results}
  \end{threeparttable}
\end{table*}

In our experiments on the EPIC-Kitchens-55 (EK-55)~\cite{Damen2018EPICKITCHENS} and EGTEA Gaze+ (EGTEA)~\cite{li2018eye} datasets, we adopt the evaluation protocol of Nagarajan et al.~\cite{nagarajan2020ego}, employing mean Average Precision (mAP) as the performance metric for this multi‑label classification task. Specifically, for each dataset, INSIGHT is provided with the first \(P\%\) of video frames (\(P \in \{25,50,75\}\)) as input and is tasked with predicting the actions occurring in the remaining \((100 - P)\%\) frames. We then compute the mAP on the validation set for all actions (All), frequently occurring actions (Freq), and rare actions (Rare). Finally, to obtain the values reported in this paper, we average the mAP scores across the three observation horizons for each action category. Results are shown in Table~\ref{tab:lta_results}.

\begin{figure}[t]
  \centering
  \includegraphics[width=\columnwidth]{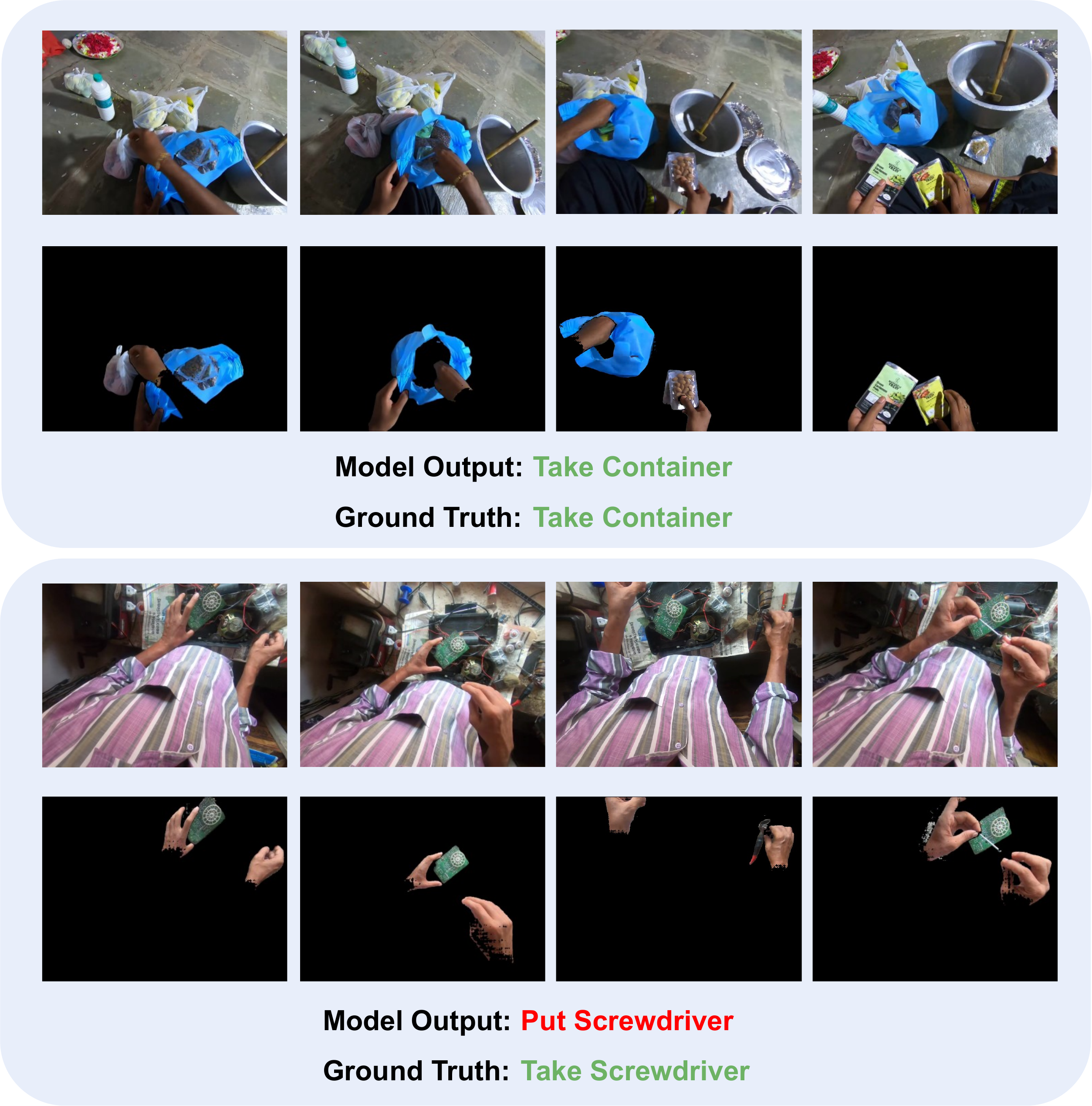}
  \caption{Successful and failed cases of Hand-Object Semantic Action Recognition module.}
  \label{fig:AR_example}
\end{figure}

\subsection{Case Study}


Figure~\ref{fig:AR_example} illustrates both successful and failed examples from our hand-object semantic action recognition module. Our module successfully identifies Hand-Object Interaction (HOI) regions and segments them correctly. In the failed case, the module confuses the inverse verb pair ``put'' and ``take'', predicting ``put'' when the ground truth is ``take''. This error suggests a weakness in capturing short-range state transitions and action polarity, which are essential to disambiguate these verbs. To address this limitation, future work could adopt a window-based observation scheme to capture cues immediately before and after the action, and integrate motion‑direction priors into the recognition process to better distinguish inverse actions.


\section{Limitations}

Two directions are particularly promising. \textbf{(i) Enhanced egocentric visual grounding.} Although our HOI module already improves recognition, further advances in modeling egocentric visual dynamics-e.g., adaptive spatio-temporal HOI tracking, affordance-aware object state modeling, and large-scale self-supervised pretraining on unlabeled egocentric streams-may yield higher action recognition accuracy and, by extension, sharper anticipation. \textbf{(ii) Advanced LLM fine-tuning for reasoning.} We plan to investigate more expressive policy optimization and structure-aware objectives (e.g., hierarchical intention decomposition, uncertainty-aware rewards, and memory-augmented reasoning) to better align the LLM’s internal cognitive traces with task semantics. We believe progress along these axes will narrow the gap between machine and human foresight in complex egocentric environments.


\bibliography{aaai2026}

\end{document}